\documentclass{article}

\usepackage{arxiv}
\usepackage[utf8]{inputenc}
\usepackage[T1]{fontenc}
\usepackage{hyperref}
\usepackage{cleveref}
\usepackage{url}
\usepackage{booktabs}
\usepackage{amsfonts}
\usepackage{nicefrac}
\usepackage{microtype}
\usepackage{lipsum}
\usepackage{amsmath}
\usepackage{graphicx}
\usepackage{subcaption}
\usepackage{enumitem}

\graphicspath{{./img/}}

\title{Credit scoring for micro-loans}

\author{
    Nikolay Dubina, Dasom Kang, Alex Suh\\
    Balancehero\\
    \texttt{\{nikolay.dubina,jules.kang,alex\}@balancehero.com}
}

\begin{document}
\maketitle

\begin{abstract}
    Credit Scores are ubiquitous and instrumental for loan providers and regulators. In this paper we showcase how micro-loan credit system can be developed in real setting. We show what challenges arise and discuss solutions. Particularly, we are concerned about model interpretability and data quality. In the final section, we introduce semi-supervised algorithm that aids model development and evaluate its performance.
\end{abstract}


\section{Introduction}
Credit score is widely adopted tool in financial industry that has been repeatedly proving its effectiveness in minimizing risk. The primary function of a credit score is to assist loan providers in estimating risk while adjusting eligibility and fees for their products. Ultimately, it is only one of the factors considered in making final decision, although often it is one of the most important ones.

Credit score is a numeric value that reflects probability that individual will default on particular kind of loan product. In popular scores, such as FICO and CIBIL, higher value specifies smaller likelihood of the default. Commonly, defaulted loan is defined as condition when individual did not repay loan in full before expected repayment date. Repayment time varies from weeks (e.g. micro-loans) to years (e.g. real estate). Alternative interpretations are also possible. In some cases, it may be beneficial to estimate time until full repayment (if most loans are repaid) or likelihood of eventual repayment (if most loans are not repaid). In this paper we are focused at estimating probability,

$$
    p(y|\theta_{loan}, \theta_{individual})
$$

$$
y = 
\begin{cases}
    +1  & \text{if loan defaulted (not repaid in full by expected date)} \\
    -1  & \text{otherwise} \\
\end{cases}
$$

\begin{samepage}
Where,
\begin{itemize}
    \item[] $\theta_{individual}$ - parameters describing individual at time of taking loan (e.g. income level)
    \item[] $\theta_{loan}$ - parameters describing loan (e.g. principal amount, duration)
\end{itemize}
\end{samepage}

Traditional loans deal with large principal amounts and long repayment times. However, certain customers can greatly benefit from smaller principals and shorter repayment times. Example products include utility bills and subscription services, such as Balancehero \cite{balancehero}. In fact, micro-loans reduce financial vulnerability and empower underprivileged groups, and therefore they contribute to alleviation of poverty and inequality. They are particularly effective in developing countries such as India and Bangladesh, with most notable successful implementation done by a Nobel Peace Prize recipient, Grameen Bank \cite{credit-micro, credit-micro-india-1}. Micro-loans have advantages for practitioners too. Short repayment times and small principal amounts allow to collect labeled data faster and have wider population coverage. Lastly, micro-loans benefit from regulation perspective. In this paper we are focusing on micro-loans, although some points may be applicable to traditional loans as well.

Credit score directly affects profitability of business. It is important to optimize for traditional classification metrics such as accuracy or AUC ROC. At the same time, profound impact of credit scores on society imposes additional responsibility on its use and development. Particularly, credit score has to be interpretable, stable with time and fluctuations in $\theta_{individual}$ and $\theta_{loan}$. Its decisions should be supplied with appropriate recourse, in case its decisions are being contested later. Next, classifiers should not be biased against protected groups. In fact, many governments legally oblige financial institutions on satisfying these requirements. We address this issues in \Cref{sec:interpret}.

\section{Dataset}
\label{sec:data}

Data collection for this problem is not an easy task. In contrast to other domains, such as Computer Vision or Natural Language Processing, there is no fairly fast and cheap way of collecting labeled samples. Instead, primary source of labeled data comes from pilot programs and test studies spawning over periods of months, where small fraction of population is exposed to loan product. Number of labeled samples is limited and it takes long time to produce them. In fact, it is not uncommon for credit system in traditional loans to spend years for data collection, which leads to operational challenges and numerous problems with consistency of features and labels \cite{practical, practical-comment}.

For our first model we had only 2,200 samples, meanwhile inference dataset contained more than hundreds of thousands. There are numerous studies showing that it is possible to successfully train classifiers for credit scoring with small amount of data \cite{credit1,credit2,credit3,credit4}. However, it is not clear how well these models translate to real production setting. Moreover, they are not strictly applicable to our case since we are not relying on typical features for credit assessment. Rather we are using sparse indirect features. Therefore, small size of labeled data was particularly acute. Once product was released, number of labeled samples quickly grew and aforementioned issues seized to be critical.

\subsection{Train-Test split}
In order to reduce effect of look-ahead bias, we select data for training only up to holdout date. This date is selected such that 20\% of data belongs to test set. Randomly sampling fraction for test set without holdout date, consistently lead to higher performance. To make tests harder we avoided it. Moreover, for initial training data, using both time holdout and random fraction resulted in either too short duration of time holdout or insufficient amount of training data.

\subsection{Labels}
As highlighted by Hand \cite{practical}, in production environment labels are prone to \textit{concept drift} --- a phenomenon where labels during training and inference do not match. To address this, we are defining labels in simplest way possible that is also acceptable by downstream users\footnote{By model "users" we mean loan providers and regulatory bodies, as opposed to "individuals" who are taking loans.}. Next, whenever definition has changed, we are retraining model on historical data and adjusting regular retraining and inference cycles.

Second, true miss-classification cost at inference time is not known at training time. However, during model development we incorporating feedback from downstream users on which groups of loans ($\theta_{individual}$, $\theta_{loan}$) are expected to be the most important. We try to match these groups as close as possible at training time. Effectively, we try to maintain equal miss-classification cost by doing similar sampling. That being said, there is still discrepancy between what model has been optimized for and how it is being used in practice.

\subsection{Features}
Our dataset is comprised of in-app user activity. All features are produced by in-app event logs\footnote{All information is obtained with the consent of the app users ("individuals").}. This includes engagement, in-app purchases, profile information, geographical data, network activity, device information, in-app social activity. Geographical features include information on distance to the closest city and village center, regularity of most common locations, distance travelled. When possible, categorical features are cross-referenced with open databases to reduce dimensionality and provide meaningful comparison metrics (e.g. "village population size"). All events are aggregated with multiple granularity and aggregation methods. For example, we break down “successful purchase” event by week, day of week, hour of day and aggregate with mean, max and min. In total, this results in 1800 sparse features with one-hot-encoding (1300 numeric, 40 categorical). In order to prevent feature leaks, features for each sample include data only up to a date of issuing loan product.

Next, it is common in practice that features during training and inference are not from the same distribution. There are a couple of reasons to that.

\paragraph{Population drift}
First, feature distribution is non-stationary \cite{practical, practical-comment}. To name a few, upstream event logs are changing with app versions, user interface updates changes and with it meaning behind the same events changes too, external conditions including national holidays, seasons and state of economy influence human responses. To address easiest of them, we define features such that their meaning does not with new app versions. For all other drifts, we regularly retrain model and keep track of what was the major external conditions in order to adjust our expectations on model's performance.

\paragraph{Reject inference}
Another fundamental issue, particularly important in credit scoring, is bias in labeled data due to nature of operation \cite{practical}. Labeled data, that is coming as a result of selecting loans based on credit score, is fundamentally biased by model predictions. If we are to solely rely on it, we would discard labeled data for loans that are expected to be bad. Instead, labeled samples should come from inference distribution. To reduce this effect, we are avoiding use of credit score for small fraction of total population. It is worth noting, if this approach is too costly, model based methods can be used instead to account for this bias \cite{credit-reject-inference, credit-reject-inference-dnn}.

\paragraph{Operational setting}
Lastly, inference dataset is not static. Downstream users are relying on credit scores to assist their decision making for different groups which they change often. Thus, we track separate overall performance metrics (e.g. accuracy) for deployed model and compare them to historical top-line values.

\section{Interpretability}
\label{sec:interpret}

Main goals for credit score is to aid decisions for its users and to help regulate on how these decisions are being made. In practice, it is insufficient to provide just the score itself (i.e. result of \textit{black-box} models) without supplementary information about how this decision has been made. Additionally, good interpretability helps developers to make improvements, debug and maintain models. Although, there is no consensus on what qualifies as interpretable or not, a great way to approach this notion, as summarized by Lipton \cite{interpret}, is to consider desired properties of model and its post-hoc use.

\subsection{Model properties}
Algorithm training procedure and computation at inference time should be clear --- \textit{algorithmic transparency}. Its stronger extension, \textit{simultability}, requires full replication of model results by humans. Reasonably regularized common algorithm including random forests, logistic regression, SVM (linear kernel) --- all satisfy this condition. Moreover, there is an argument that compact Neural Networks follow this property as well \cite{interpret}.

Model parameters and computation steps have to be individually explainable --- \textit{decomposability}. To achieve this, we are selecting only simple enough features with intuitive explanation. Moreover, computation for these features and their groups can be described in plain language. Other approaches that aid this goal are also possible \cite{credit-interpret-copy}.

\subsubsection{Operational requirements}
Successful utilization of credit score by downstream users imposes extra requirements. Particularly, probability score should be stable --- \textit{sensitivity}. This manifests itself in multiple ways. Model decisions should be insensitive to noise (e.g. errors in measurements). It should have small rate of change locally (e.g. small valid changes in activity) \cite{practical-comment}. Similar individuals need to have similar credit scores. Individuals should maintain similar credit score over time, unless their $\theta_{individual}$ changes significantly (i.e. temporal consistency). In broader context, study of local properties of decision boundary is of particular interest in other domains where cost of errors is high, such as adversarial samples in Computer Vision \cite{semi-adversarial}. In \Cref{sec:unlabeled} we are introducing simple algorithm that allows to test and enforce this requirement for tabular data.

\subsubsection{Regulatory requirements}
Depending on jurisdiction, regulatory bodies have additional set of required properties. First, it is common to have a restriction on how much of historical activity is allowed for decision making --- \textit{forgetfulness}. For example, in U.S. Medicaid look-back period is limited to 5 years \cite{credit-us-medicaid} and according to U.S. Federal Reserve System credit scores look-back time goes up to 10 years \cite{credit-us-gudeline}. In context of micro-loans, it can be required that at most three months of most recent "purchase history" can be used. The easiest solution to this is to consider smallest allowed look-back for individuals’ information and use it for all features. This fits micro-loans well, since most recent activity is often the most important one.

Next, in case of model's decisions are being contested, models should provide actionable \textit{recourse} \cite{credit-recourse}. Recourse is a set of actions individual needs to take in order to change credit scoring decision. First, model should guarantee that it is feasible for individuals to change it. Next, it should provide list of such actions for particular case. In turn, these actions lead to updates in mutable features in $\theta_{individual}$. In our case, almost all data is coming from recent user activity and therefore it is mutable. Lastly, as described in \cite{credit-recourse}, the cost of recourse varies between groups of individuals. Optimizing for appropriate balance is subject of further investigations.

These requirements are not exhaustive. There are other important issues that regulators can be focused on, such as \textit{equal opportunity}, which is addressed at length in \cite{credit-fairness}.

\subsection{Post-hoc interpretation}

\begin{samepage}
\begin{figure}
    \centering
    \begin{subfigure}{.32\textwidth}
        \centering
        \includegraphics[width=.8\linewidth]{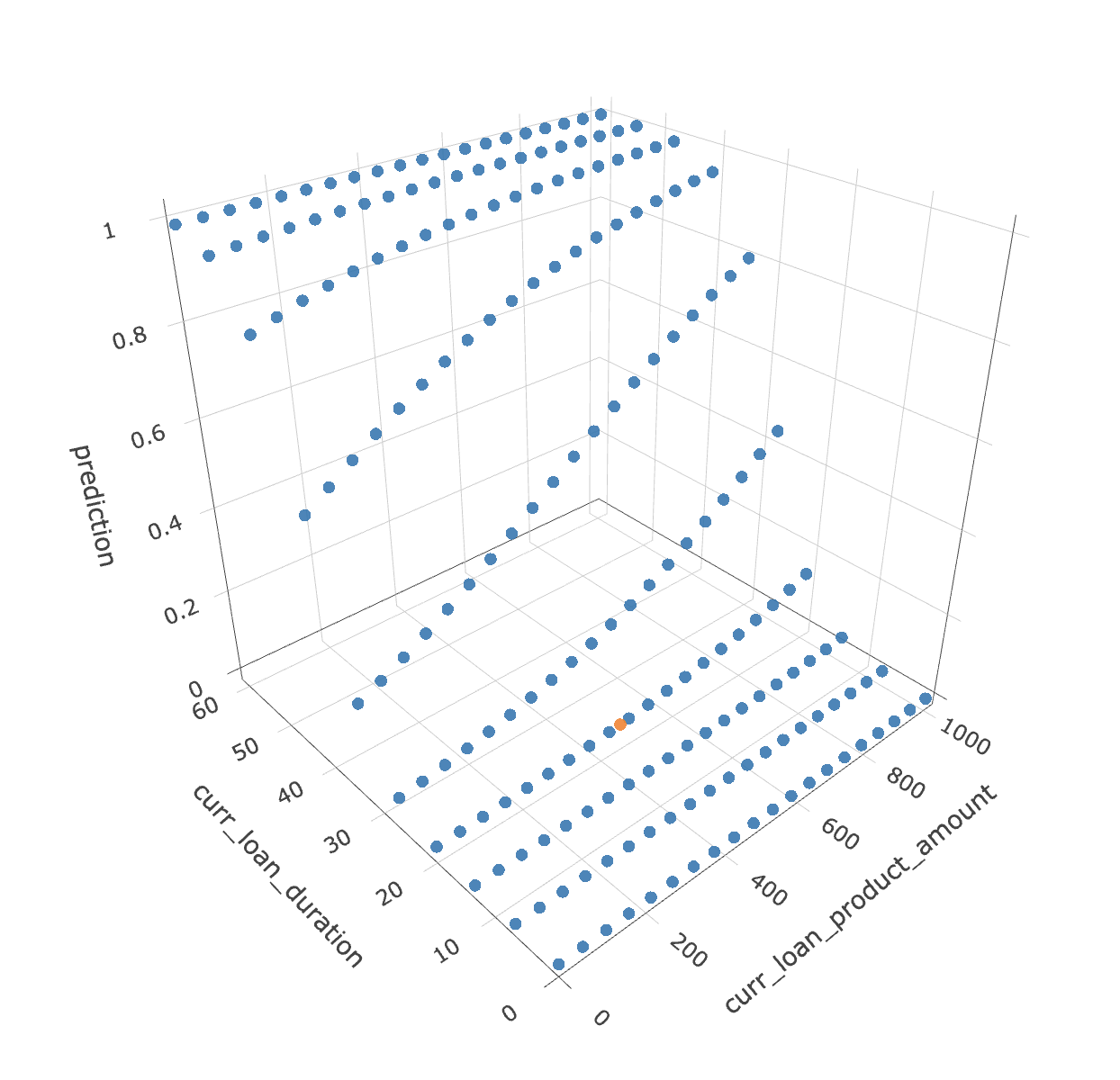}
        \caption{Logistic Regression}
        \label{fig:posthoc-admissible-lin}
    \end{subfigure}
    \begin{subfigure}{.32\textwidth}
        \centering
        \includegraphics[width=.8\linewidth]{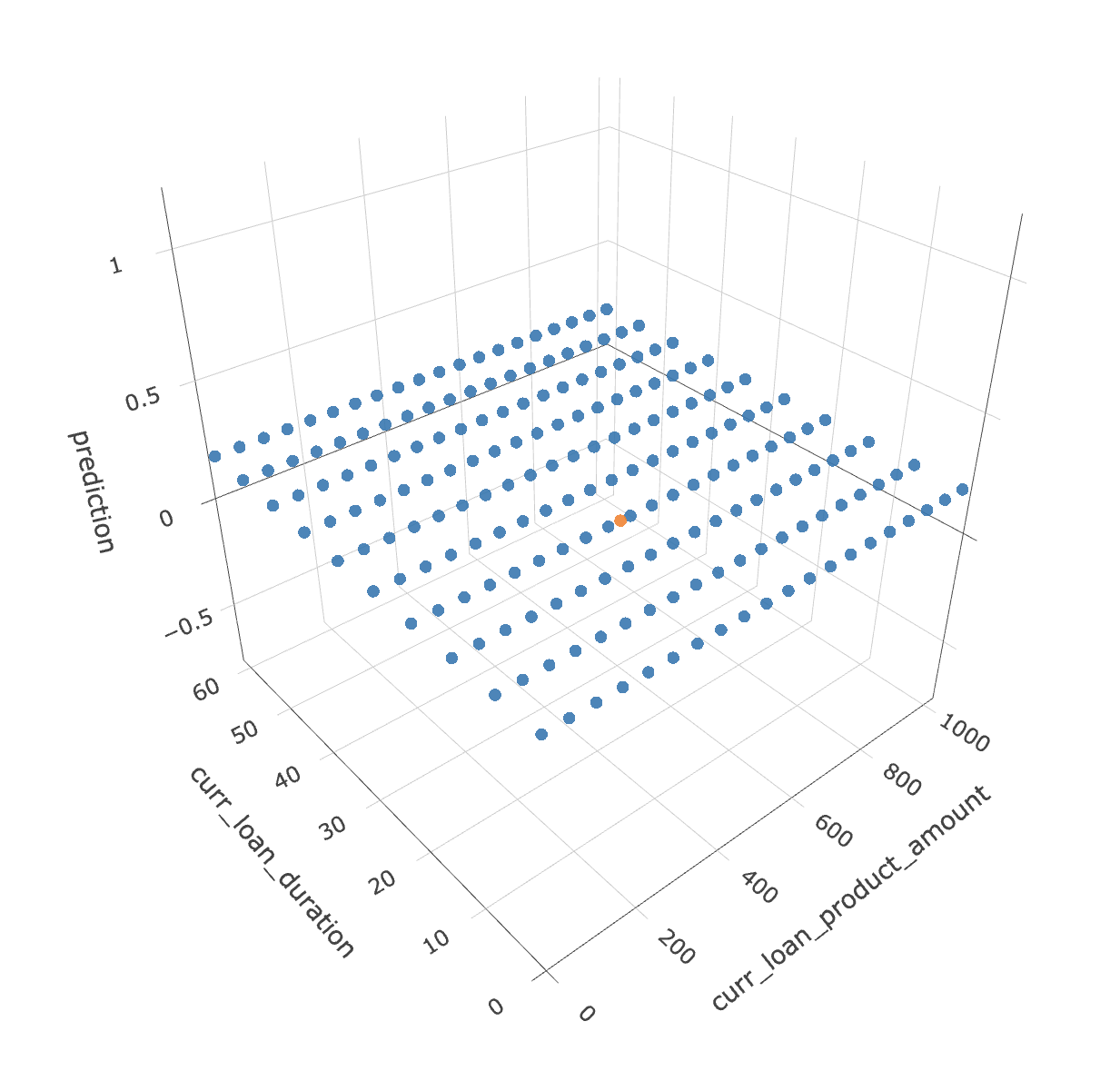}
        \caption{Random Forest}
        \label{fig:posthoc-admissible-xgboost}
    \end{subfigure}
    \begin{subfigure}{.32\textwidth}
        \centering
        \includegraphics[width=.8\linewidth]{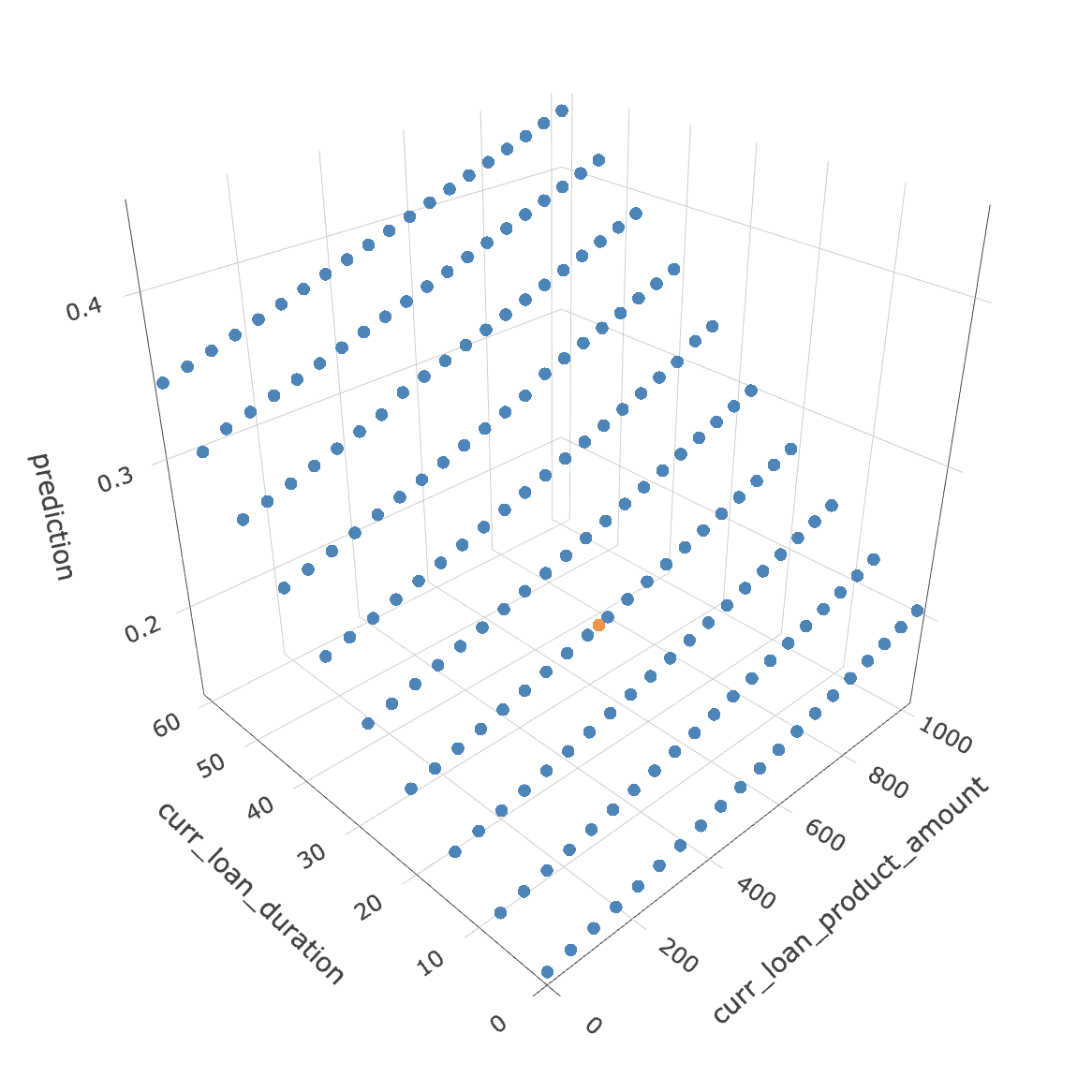}
        \caption{SVM (RBF kernel)}
        \label{fig:posthoc-admissible-lsvm}
    \end{subfigure}
    \caption{Example of final probability with changes of two in input features, while other features being constant, for selected sample. Decision boundary for random forest is always a mixture of parallel surfaces. However, node splits are cane be performed on other features in a way that reflects proportional changes in features of interest (correlated features). This makes analyze more complex.}
    \label{fig:posthoc-admissible}
\end{figure}
\begin{figure}
    \centering
    \begin{subfigure}{.32\textwidth}
        \centering
        \includegraphics[width=.8\linewidth]{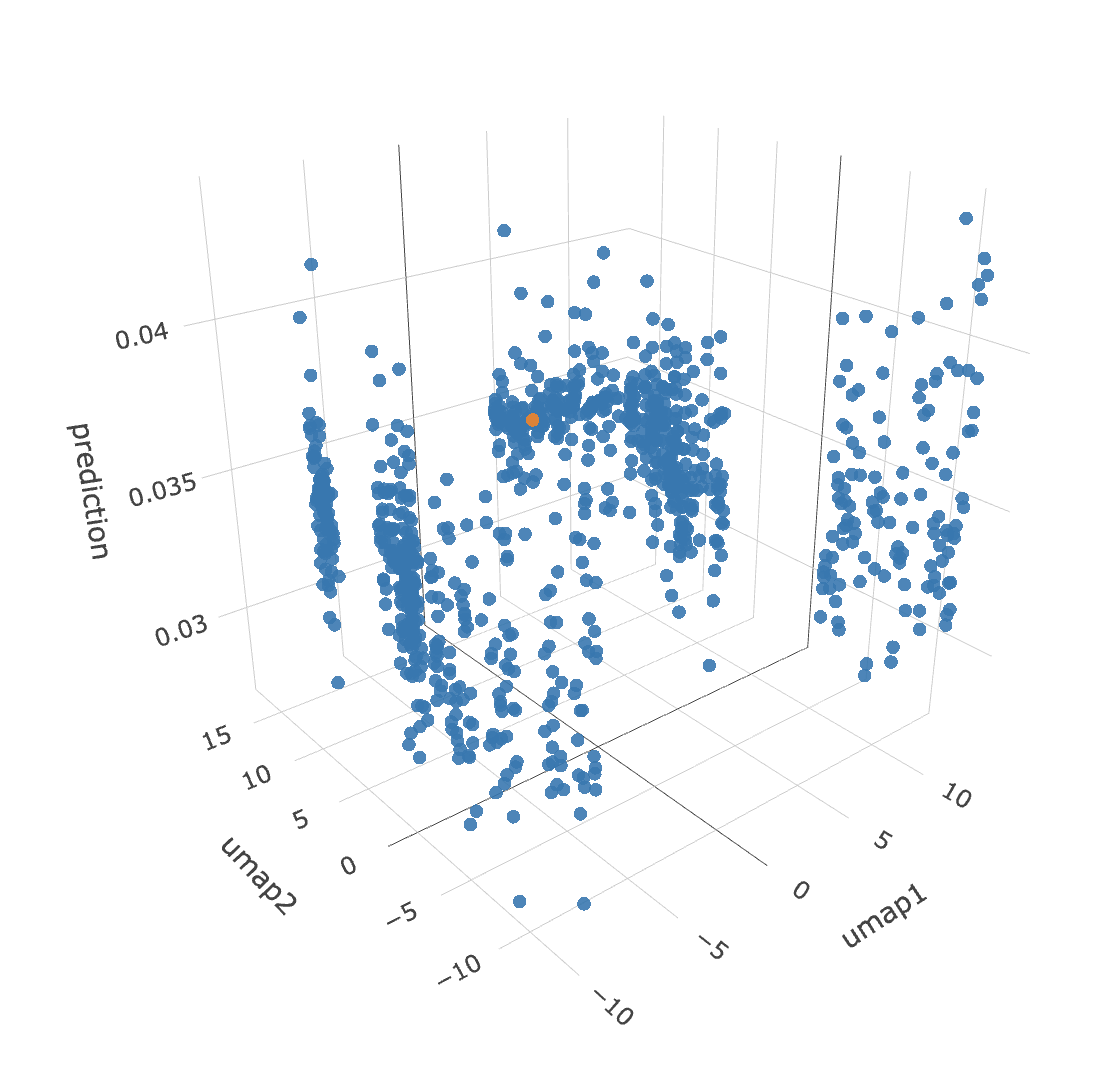}
        \caption{Logistic Regression}
        \label{fig:posthoc-embedding-lin}
    \end{subfigure}
    \begin{subfigure}{.32\textwidth}
        \centering
        \includegraphics[width=.8\linewidth]{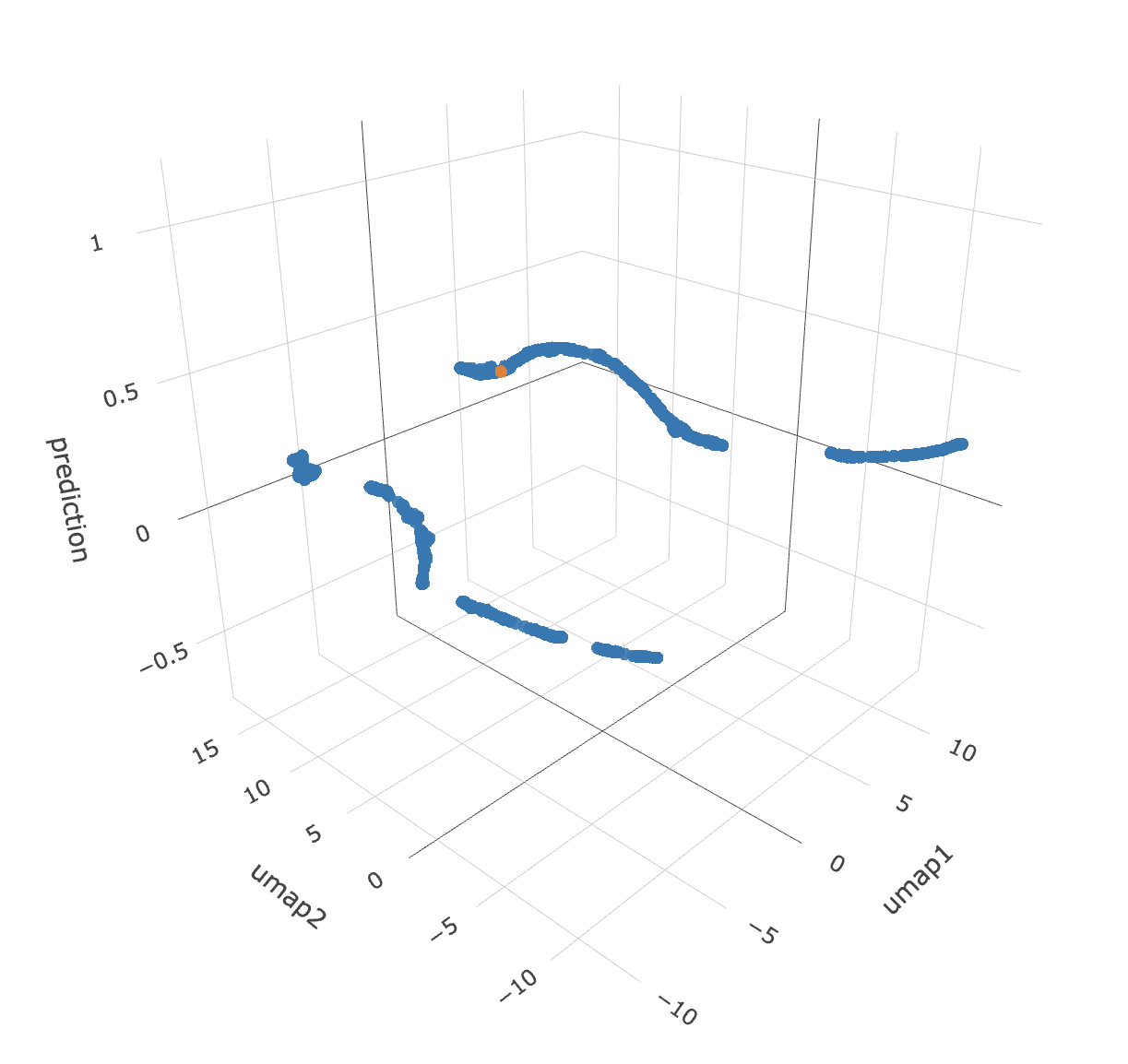}
        \caption{Random Forest}
        \label{fig:posthoc-embedding-xgboost}
    \end{subfigure}
    \begin{subfigure}{.32\textwidth}
        \centering
        \includegraphics[width=.8\linewidth]{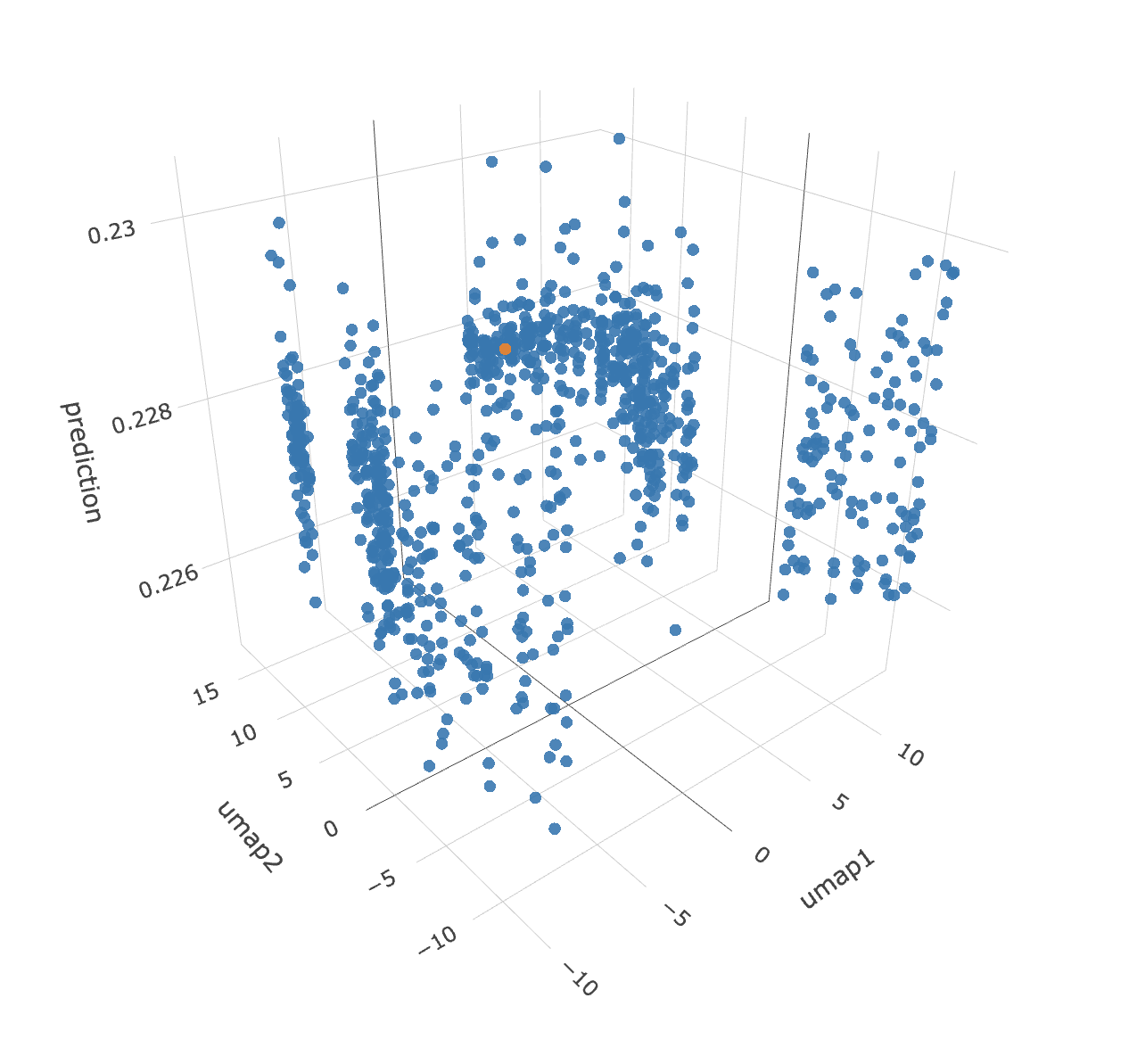}
        \caption{SVM (RBF kernel)}
        \label{fig:posthoc-embedding-svm}
    \end{subfigure}
    \caption{Example of final probability for changes in selected features within fixed total similarity, while other features being constant, for selected sample. Similarly to \Cref{fig:posthoc-admissible}, random forest keeps piece-wise predictions. Meanwhile, other algorithms have different spread in different regions based on sample similarity.}
    \label{fig:posthoc-embedding}
\end{figure}
\end{samepage}

Effective method of insuring model interpretation is visualization of its properties. With this approach \textit{algorithmic transparency} and \textit{decomposability} can be illustrated by providing list of selected features and how they contribute to decisions (e.g. decision tree plot).

Properties of \textit{sensitivity} and \textit{recourse} can be visualized with help of local explanations \cite{interpret, interpret-lin}. Particularly, we can show how model predictions change for selected individual with changes in features. One example is to show how changes in two selected features, others being fixed, influences model prediction (\Cref{fig:posthoc-admissible}). Similarly, we can do illustrative test for susceptibility to noise or existence of recourse for selected samples (\Cref{fig:posthoc-embedding}). To do that we vary selected features within fixed similarity (\Cref{sec:unlabeled}) and compute UMAP\footnote{Uniform Manifold Approximation and Projection for Dimension Reduction (UMAP) is fast embedding that preserves global structure \cite{umap}.} embedding for dimensionality reduction. If similarity threshold is sufficiently large, then we can observe existence of points that cross decision boundary, in other words --- existence of recourse. Moreover, given appropriate kernel and features, each such point is supplied with simple explanation on how to change features to achieve it. It is also shown in LIME how to locally approximate arbitrary decision boundary with linear models \cite{interpret-lin}.

Furthermore, model users and regulators might require extra post-hoc information on case-by-case basis. For example, if credit score is being contested, appropriate list of \textit{recourse} actions (and \textit{flipset} for $\theta_{individual}$) should be provided. For example, authors of \cite{credit-recourse} show how such lists can be attained for linear models and necessary conditions for their existence.

\section{Baseline classifiers}
We have used a variety of standard algorithms, including logistic regression, random forests (XGBoost \cite{xgboost}), SVM, Neural Networks\footnote{Multi-layer perceptrons with batch normalization and tanh activation. Sigmoid activation consistently lead to inferior results.}. They repeatedly show top-line performance on many tasks on tabular data, including credit scoring \cite{credit1, credit2, credit4, credit5}. For each algorithm we varied regularization parameters and subsets of selected features to achieve highest AUC ROC and reasonably simple model (e.g. small decision trees, small number of parameters for linear models).

As mentioned earlier in \Cref{sec:data}, our training and test data is limited. To combat overfitting, we conduct hyperparameter search (i.e. $\alpha$ and $\beta$ for XGBoost, amount of $l_1$ and $l_2$ regularization for logistic regression). Next, we varied selected features by manual review. We also cross-referenced selected features based on their importance in other models that have enough labeled data and high historic accuracy.

\section{Finding similar samples in unlabeled data}
\label{sec:unlabeled}

In order to check how stable model predictions are we are requiring that similar samples should have same prediction. One approach would be to sample randomly in neighbourhood of labeled data, however major shortcoming of this approach is that it is meaningful only for small distance around labeled samples in feature space. Thus, to make similar samples more realistic we would like to use unlabeled data. There are a number of semi-supervised algorithms developed with that aim \cite{semi-fuzzy, semi-cluster, semi-post-label}. In general, semi-supervised algorithms exploit at least one of the assumptions about data, 

\begin{itemize}
    \item \textit{Smoothness} --- labels should not be sensitive to small changes in features.
    \item \textit{Low density separation} --- labeled and unlabeled data coming from the same distribution and forms clusters separated by low-density regions. Decision boundary should go through these regions.
    \item \textit{Manifold} --- samples from labeled and unlabeled data come from the same distribution and lie on low dimensional manifolds. Labels for samples on same manifolds should be the same of change smoothly.
\end{itemize}

\begin{figure}
    \centering
    \begin{subfigure}{.32\textwidth}
        \centering
        \includegraphics[width=\linewidth]{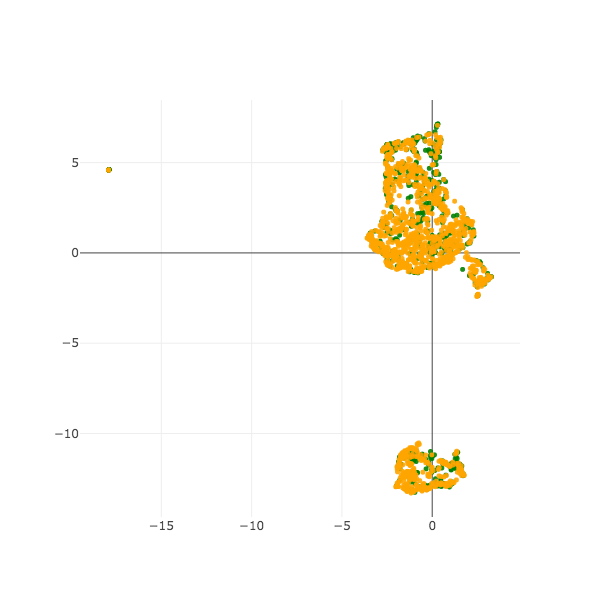}
        \caption{Labeled samples}
        \label{fig:alg-labeled}
    \end{subfigure}
    \begin{subfigure}{.32\textwidth}
        \centering
        \includegraphics[width=\linewidth]{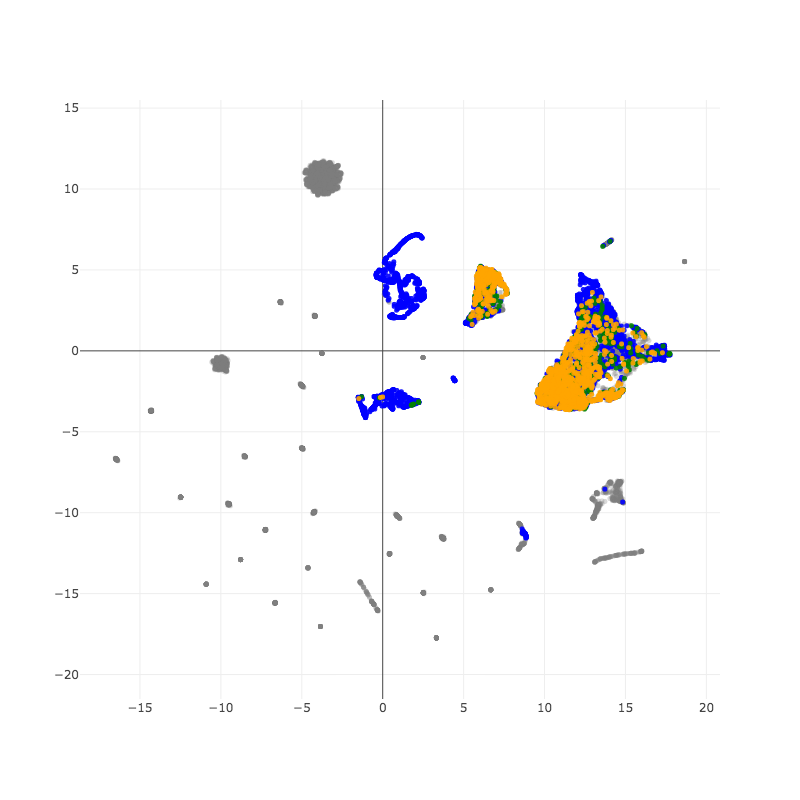}
        \caption{Labeled and similar samples}
        \label{fig:alg-similar-labeled}
    \end{subfigure}
    \begin{subfigure}{.32\textwidth}
        \centering
        \includegraphics[width=\linewidth]{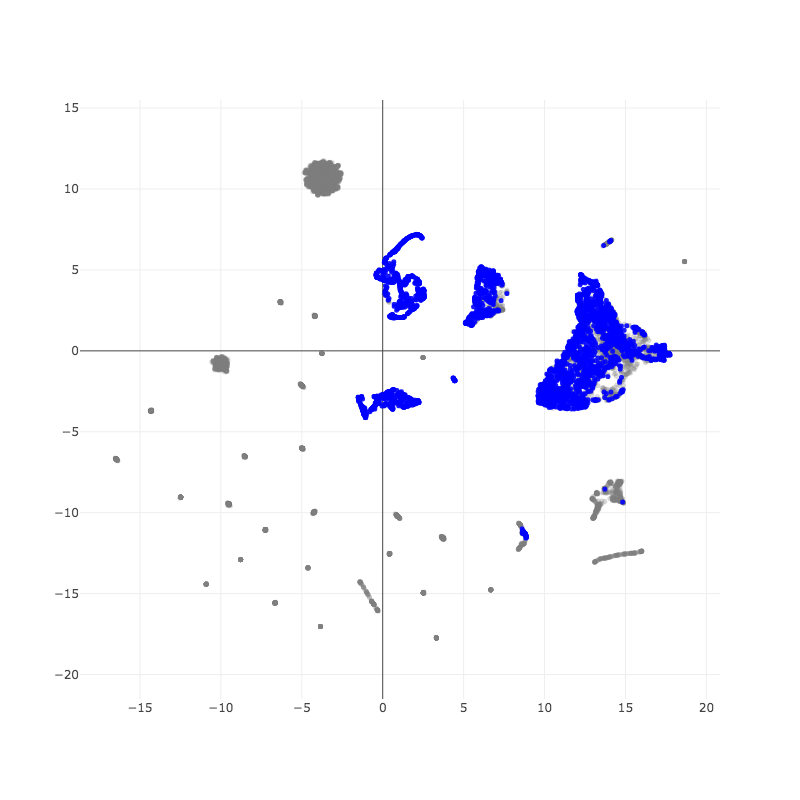}
        \caption{Similar samples}
        \label{fig:alg-similar}
    \end{subfigure}
    \begin{subfigure}{.32\textwidth}
        \centering
        \includegraphics[width=\linewidth]{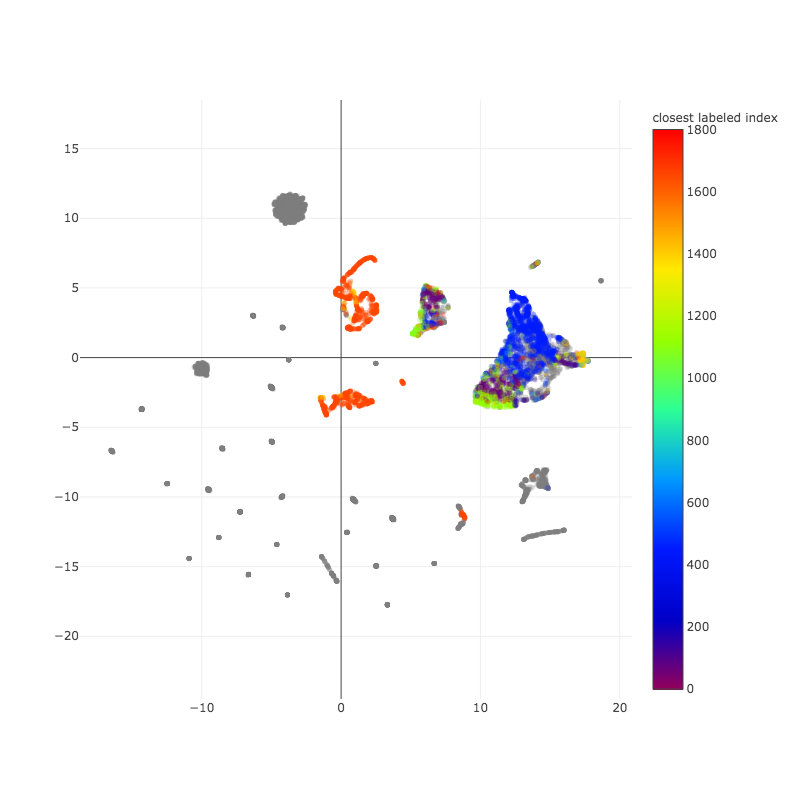}
        \caption{Index of closest sample}
        \label{fig:alg-closest}
    \end{subfigure}
    \begin{subfigure}{.32\textwidth}
        \centering
        \includegraphics[width=\linewidth]{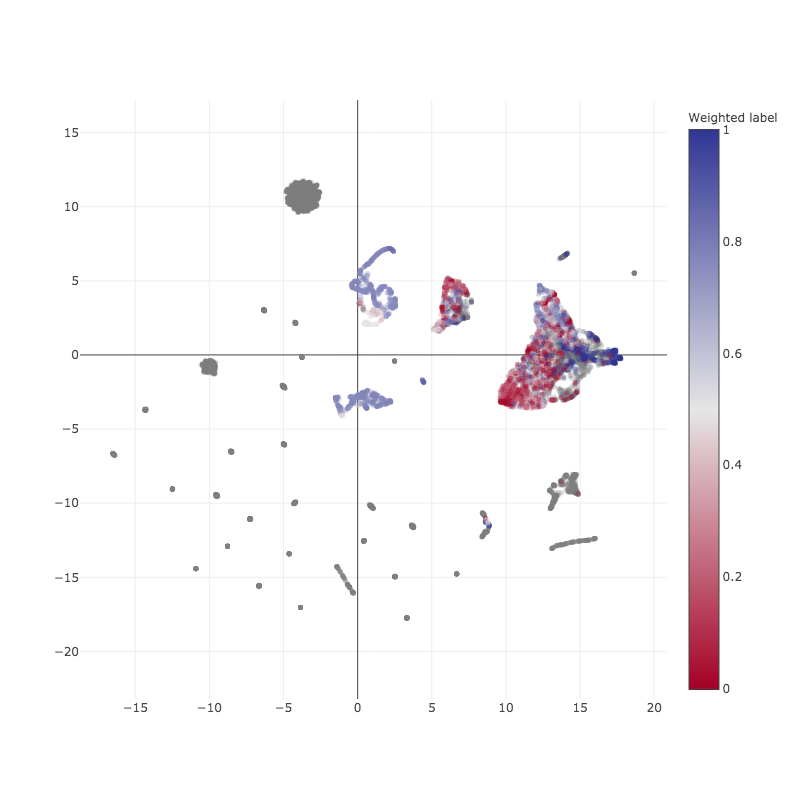}
        \caption{Label estimation for similar samples}
        \label{fig:alg-weighted}
    \end{subfigure}
    \begin{subfigure}{.32\textwidth}
        \centering
        \includegraphics[width=\linewidth]{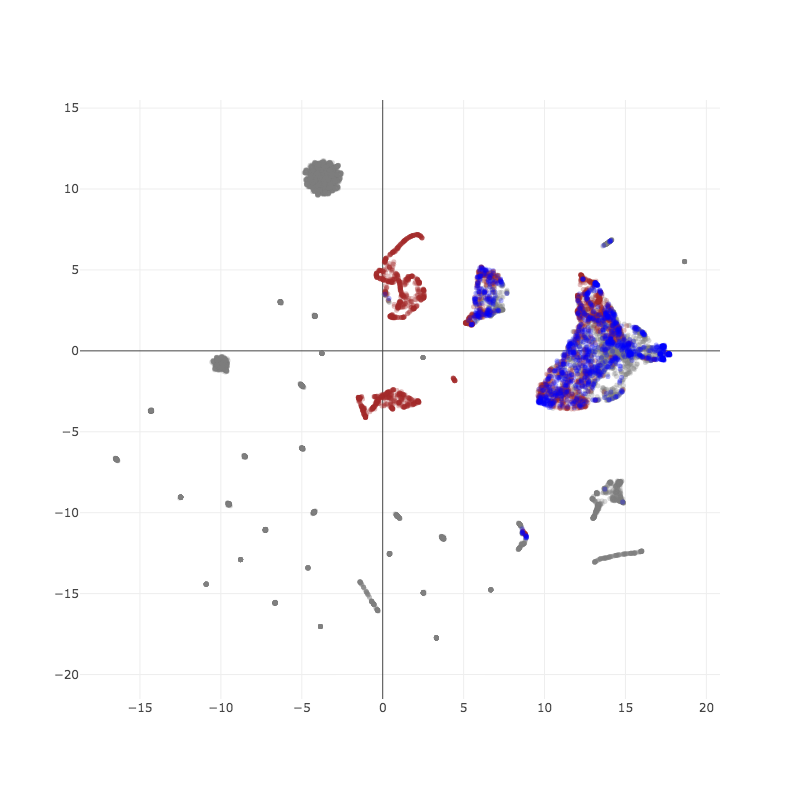}
        \caption{Confident similar samples}
        \label{fig:alg-confident}
    \end{subfigure}
    \caption{UMAP embedding of labeled and unlabeled data using all features.}
    \label{fig:alg}
\end{figure}

Labeled data is not sampled at random (MNAR) from unlabeled data. First, individuals whitelisted for pilot programs have been sampled not at random and the exact methodology is not accessible. Moreover, only small fraction of the batch participated. This leads to vastly different feature and label distributions. As reported by other studies, such effect can be detrimental to already low efficiency of semi-supervised methods \cite{semi-comp-1}. Next, unlabeled data does have clusters with low density separation, but they do not have clear majority of labeled samples (\Cref{fig:alg-labeled}). Thus, low density separation algorithms (e.g. TSVM family \cite{tsvm, tsvm1}, post-labeling family \cite{semi-post-label}) are not applicable. As we observed, unlabeled samples can not be efficiently reduced to low dimension manifolds (\Cref{fig:alg-similar}). Hence, manifold algorithms (e.g. label propagation) are not applicable either. This is not new. Limited applicability of semi-supervised algorithms has been discussed in \cite{semi-crit, semi-cluster, semi-comp}. Low performance of algorithms relying on either of mentioned approaches has been verified empirically.

Therefore, we are only enforcing smoothness assumption in terms of similar labels from unlabeled data. By following algorithm described in \Cref{alg-descr} we are finding similar samples to training data and separately similar samples to test data. Next, these datasets can be used to boost training set and help model selection.

\subsection{Algorithm description}
\label{alg-descr}
Let labeled dataset be $D_l = \{(y_i, x_i)\}_{i=1}^N$, where $y_i \in \{-1, +1\}$ and features $x_i \in X$. Similarly, unlabeled dataset is $D_u = \{(\hat{y}_j, \hat{x}_j, x_j)\}_{j=1}^M$, where estimation of label $\hat{y}_j \in \{-1, 0, +1\}$, estimation of features missing in unlabeled samples $\hat{x}_j \in 
(X - X^{\star})$ and features present in both datasets $x_j \in X^{\star}$. Note, any classifier on $D_l$ can also make predictions on $D_u$. If algorithm can not assign label to unlabeled sample then $\hat{y}_j = 0$. Let us also define kernel function $k \colon X^{\star} \times X^{\star}  \to [0, 1]$ that represents similarity between two samples. Label estimation algorithm is as follows,

$$
w_{ij} = \mathbb{I}(k(x_i, x_j) > d) \cdot k(x_i, x_j)
$$

$$
t_j = \frac{\sum_{i=1}^{N} w_{ij} \cdot y_i}{\sum_{i=1}^{N} w_{ij}}
$$

$$
\hat{y}_{j} = 
\begin{cases}
    +1  & \text{if $t_{j} > c $} \\
    -1  & \text{if $t_{j} < -c $} \\
    0   & \text{otherwise} \\
\end{cases}
$$

$$
\hat{x}_{j} = 
\begin{cases}
    \frac{\sum_{i=1}^{N} w_{jk} \cdot x_i}{\sum_{i=1}^{N} w_{jk}} & \text{if $t_{j} > c $ or $t_{j} < -c $} \\
    \textit{null}   & \text{otherwise} \\
\end{cases}
$$

\begin{samepage}
Where,
\begin{itemize}
    \item[] $d$ - similarity threshold
    \item[] $c$ - confidence threshold
\end{itemize}
\end{samepage}

This algorithm estimates label and unknown features for unlabeled samples that are similar enough and lie in regions of feature space with clear majority of either of positive or negative labels. Selecting sufficiently similar unlabeled samples allows to correctly select samples in the same cluster as labeled data \Cref{fig:alg-similar}. Using closest similar sample for estimation is does not produce stable decision boundary \Cref{fig:alg-closest}. However, filtering confident samples \Cref{fig:alg-weighted} produces good final result \Cref{fig:alg-confident}.

Both parameters $d$ and $c$ control trade-off between purity of labels of similar samples. First, labeled samples themselves has to be sufficiently \textit{un}-similar to each other or otherwise all of them will match unlabeled data and no meaningful estimation would be possible. On the other hand, similarity threshold $d$ has to be high enough to match sufficiently many unlabeled samples. A good choice is to fix $d$ at 95\% percentile of similarity score between labeled samples. Similarly, higher value of $c$ gives higher purity but fewer label assignments and lower number of selected unlabeled samples. A good choice for $c$ is such that <5\% of unlabeled samples are matched to labeled data.

Lastly, given small number of labeled samples, this algorithm is highly parallelizeable. This allows to process large number of unlabeled samples and in turn find sufficient number highly similar and confident samples.

\subsubsection{Kernel function}

Choice of distance metric is a challenging task and greatly influences performance. In semi-supervised methods for image classification a common choice is $l_2$ metric \cite{semi-cluster, semi-adversarial}. It is also possible to  learn adaptive cluster depending metric \cite{semi-fuzzy}. Generally, choice of metric is subject to type of data at hand and the qualities that practitioners expect that data to satisfy. To simplify development, we assume that all selected features as equally important. Moreover, we consider only quantitative features\footnote{Kernel can be trivially extended with qualitative features by following procedures explained by Gower \cite{gower}}. Since not all features are comparable to each other, we instead compare their divergence to the maximum divergence across all samples. Effectively, this is general coefficient of similarity introduced by Gower \cite{gower}. Kernel function is defined as,

$$
r_k = |max_i(x_{ik}) - min_i(x_{ik})|
$$

$$
k(x_i, x_j) = \frac{1}{D} \cdot \sum_{k=1}^{D} (1 - \frac{\|x_{ik} - x_{jk}\|_{1}}{r_k})
$$

This score has intuitive interpretation. Consider samples with 10 features. Similarity of 90\% ($d = 0.9$) can be accounted by two scenarios: one feature with 10\% of maximum divergence and other 9 with 0\%; or by 0\% difference in one feature and other 9 with 1\%. This property of linear trade-off between individual features leads to convenient interpretation of interplay between features`s contributions to similarity between samples.

\subsubsection{Features selection}
Similarity between samples is very sensitive to selected features. Consider fixing similarity threshold for two samples and selecting features for similarity matching. Following factors influence selection.

\paragraph{Relevance}
We manually select features that we deem individually descriptive and appropriate in terms of percentage trade-off for similarity matching. Consider adding irrelevant feature $A$ (e.g. "name length"). First, samples who have similar important features $B$ (e.g. "amount of last order"), but different $A$, will be discarded. Next, samples with the same $A$ will increase range of possible allowed values for $B$. Both cases are undermining our initial premise that similar samples should have similar label.

\paragraph{Variance}
Low variance features increase allowed range for other features. Consider adding a feature with constant value to single effective feature, allowed range for later will double. Thus, we are keeping only features that have high coverage in labeled and unlabeled data. For example, we omit features in $\theta_{loan}$ since they are missing entirely for unlabeled data.

\paragraph{Dimensionality}
High number of features, albeit relevant, reduces allowed range for individual features for average case and increases maximum allowed value for individual features in extreme. This is implication of "curse" of dimensionality.

\paragraph{Confidence}
When number of features is too small, sample distribution will become complex and unlabeled samples will become increasingly similar to both samples for both positive and negative labels, leading to fewer confident label estimations. In contrast, with higher dimensions feature distributions become simpler and more confident matches appear. This is known as one of "blessings" of dimensionality.

\subsection{Results}
\label{result}

\begin{table}
    \centering
    \begin{tabular}{lll}
        \toprule
        Algorithm & Test data (real) & Test data (similar) \\
        \midrule
        logistic regression             & 0.62  & 0.78  \\
        logistic regression$^\star$     & 0.66  & 0.98  \\
        SVM (rbf)                       & 0.69  & 0.94  \\
        SVM (rbf)$^\star$               & 0.64  & 0.97  \\
        Random Forest                   & 0.69  & 0.92  \\
        Random Forest$^\star$           & 0.65  & 0.98  \\
        \bottomrule
    \end{tabular}
    \caption{Performance of algorithms in AUC ROC. Classifiers trained with similar samples marked with star.}
    \label{tab:performance-comparison}
\end{table}

For each algorithm we selected best performing hyper-parameters on real test set. All classifiers use the same similar training and test data. McNemar test was used to check if models are statistically significantly different \cite{compare-clf}.

First, we observe that using similar samples in training data improves top-line performance only for logistic regression, all other algorithms degrade in performance on real test set (\Cref{tab:performance-comparison}). This does not come as a surprise, as mentioned earlier in \Cref{sec:unlabeled}, semi-supervised algorithms are rarely improving top-line performance in this domain.

Next, when all classifiers are trained on labeled data only, both Random Forest and SVM (rbf) showed best performance on real dataset, however SVM (rbf) shows the best performance on \textit{similar} test set (\Cref{tab:performance-comparison}). Due to larger number of samples it leads to more confident tests for models similarity and thus can assist model selection. If model complexity (\Cref{sec:interpret}) and performance on the real test are equal, we would choose model with higher performance on \textit{similar} test set, because it satisfies requirement for stability of predictions better.

\section{Conclusion}
In this paper we described full scope of development of classifier for micro-loan credit scoring. We have showed challenges that arise in this process and how we approached them. The process of developing such systems in production is complex and there are numerous technical and conceptual challenges. By no means we imply that this is the only solution. Instead, we hope this work will contribute to broader discussion of credit scoring, model productionalizing, interpretability and semi-supervised learning.

\bibliographystyle{plain}

\end{document}